\documentclass[11pt]{article}
\PassOptionsToPackage{numbers, compress}{natbib}

\usepackage[preprint]{acl}

\usepackage{times}
\usepackage{latexsym}

\usepackage[T1]{fontenc}

\usepackage[utf8]{inputenc}

\usepackage{microtype}

\usepackage{inconsolata}

\usepackage{graphicx}

\usepackage{courier}
\usepackage{url}            
\usepackage{booktabs}       
\usepackage{amsfonts}       
\usepackage{nicefrac}       
\usepackage{arydshln} 
\usepackage{amsthm}

\usepackage{amsmath}
\usepackage{amssymb}
\usepackage{bbm}
\usepackage{threeparttable}
\usepackage[ruled,vlined,linesnumbered]{algorithm2e} 
\usepackage[most]{tcolorbox}
\usepackage{multirow}
\usepackage{makecell}
\usepackage{float}
\usepackage{stfloats}
\usepackage[table]{xcolor}
\usepackage[section]{placeins}
\usepackage[most]{tcolorbox}     
\definecolor{pink}{RGB}{180,50,120}

\usepackage[dvipsnames]{xcolor}

\newcommand{\pur}[1]{\textcolor{purple}{#1}}

\newcount\Comments  
\newcommand{\kibitz}[2]{\ifnum\Comments=1{\color{#1}{#2}}\fi}

\newcount\CommentsAdd  
\newcommand{\kibitzAdd}[2]{\ifnum\CommentsAdd=1{\color{#1}{#2}}\fi}
\definecolor{english}{rgb}{0.0, 0.5, 0.0}

\title{Confidence Composition for Multiagent Language Model Systems}

\author{
  Ali Elahi \qquad Michael J. Curry \qquad Barbara Di Eugenio \\
  University of Illinois Chicago \\
  \texttt{\{aelahi6, mjc, bdieugen\}@uic.edu}
}

\begin{document}
\maketitle

\begin{abstract}
Multiagent language model systems, such as collaborative reasoning and debate, produce multiple correlated candidate answers and confidence signals. However, these signals are usually calibrated only at the individual agent level, and provide no principled confidence estimate for the system’s final answer. We formulate this as a confidence composition problem: combining confidence across agents and reasoning stages while preserving both selective utility and probabilistic reliability. We study confidence-aware routing and log-odds pooling protocols that select among candidate answers and output a system-level confidence. Across five benchmarks, 30 heterogeneous and homogeneous model pairs, and two confidence estimators, our gated-fusion methods improve AUARC and reduce Brier score over single agent, standard debate, and selective debate baselines, while retaining competitive weighted F1-score as a correctness metric. We further show that our log-odds fusion is overconfident due to correlated intermediate signals. We propose a shared dependence discount that substantially improves reliability while preserving predictions. 

\end{abstract}

\section{Introduction}

Confidence estimation has become prominent throughout NLP tasks by facilitating selective prediction, human oversight, self-correction~\cite{yang2025confidence}, self-consistency~\cite{taubenfeld2025confidence}, routing~\cite{lin2025enhancing}, and reasoning \cite{razghandi2025cer}. However, existing confidence estimators are designed for a single model and a single response, and not for the final response of a collaborative process~\citep{ling2024uncertainty,tian2023just,lyu2025consistency}. This limitation is increasingly important as language model systems rely on multiple models and reasoning stages, producing several intermediate answers and confidence signals before aggregating them into a final response. In this work, we study the composition of confidence in multiagent settings. 

We use a two-agent debate \cite{du2024improving, Liang2024encouraging} as an environment to study confidence composition. Debate creates a setting with multiple correlated answer streams and confidence signals, in which agents may revise their answers through interactions. Prior work on multiagent debate has used confidence to trigger debate or resolve disagreements \citep{lin2025enhancing, fan2026imad, bai2024confidencecal, eo2025debate}, but has not formulated a final confidence for the multiagent system. They do not specify how confidence should be assigned in cases such as when agents agree, disagree, revise their answers after interaction, or produce several intermediate outputs from which the system selects its final answer. Also, reporting the confidence of a single agent ignores evidence from the rest of the multiagent process.

We study two methods to compose confidence. The first is Weighted Stream Voting (WSV), a soft-aggregation protocol that lets all answer-confidence streams vote for candidate labels after validation-fitted calibration and weighting, making the confidences more comparable. The second is Confidence-Gated Fusion (CGF), which first determines, for each agent, whether its post-debate answer is sufficiently confident to replace its zero-shot answer, and then combines the surviving streams in log-odds space. We also consider an asymmetric variant, A-CGF, which triggers one-sided persuasion when the agents' initial confidences are highly imbalanced. All calibration functions and the aggregation parameters are optimized on a validation set.

We evaluate across five benchmarks, 30 homogeneous and heterogeneous model pairs, and two qualitatively different confidence estimators, sequence probability and verbalized self-report. CGF and A-CGF improve system-level confidence discrimination over zero-shot, standard debate, and debate-on-disagreement baselines in all twelve comparisons ($p_{\mathrm{Holm}}\le .0016$). WSV is also competitive, particularly on MMLU. The gains hold under temperature-scaled Brier score in eleven of twelve comparisons, showing that the methods improve the quality of the reported probabilities rather than merely producing more extreme confidence values. Task accuracy remains competitive with respect to standard debate and confidence-based debate baselines (ConfMAD and \cite{lin2025enhancing}, DebUnc \cite{yoffe2025debunc}), and our protocols avoid the large F1 losses that unrestricted debate incurs on the more ambiguous tasks.

Finally, we analyze miscalibration from log-odds fusion. It assumes that the surviving confidence streams provide independent evidence, but the interaction between agents makes their signals dependent. We show that this dependence mainly affects calibration rather than discrimination, and a validation-fitted shared discount corrects the over-counting of evidence and improves ECE and Brier score while preserving ranking. Additionally, our study shows that both gating and fusion contribute meaningfully, whereas each alone yields substantially lower AUARC and higher t-Brier. This analysis shows that WSV is a strong composition protocol, while weighted voting on two zero-shot streams is not.

\section{Related Work}
Combining several probability estimates into one prediction is a classical problem in statistics. Opinion pooling studies how beliefs from multiple sources should be aggregated into a single probability distribution, with common approaches including linear pooling, which averages probabilities, and logarithmic pooling, which combines probabilities in log space \cite{genest1986combining, dietrich2017probabilistic}. Similar ideas appear in ensemble learning, where model predictions are combined through voting, probability averaging, stacking, or product-of-experts rules to improve robustness or predictive quality \cite{wolpert1992stacked, hinton2002training}. These methods differ in the assumptions they make about the sources being combined. Simple averaging returns a compromise between the input probabilities, while logarithmic or product-based rules increase confidence when multiple sources support the same answer.

\subsection{Confidence Estimation and Quality}
Confidence estimation is widely used in NLP as a proxy for correctness. It is often framed as the probability of a statement generated by a language model being factual ($P(True)$), or the probability of a model holding the information for an input query ($P(I Know)$) \cite{mahaut2024factual}. Existing LLM confidence estimators fall into three broad families. Latent-information methods use quantities such as token log-probabilities, sequence likelihood, entropy, or other distributional statistics \citep{ling2024uncertainty,bakman2024mars}. Verbalization methods ask the model to report a numerical confidence or likelihood in natural language \citep{tian2023just,xiong2023verb}. Consistency-based methods estimate uncertainty from agreement across multiple sampled generations \citep{lyu2025consistency}. These methods are defined and studied for a single model output.

A confidence signal is useful when it matches empirical correctness (Calibrated) and discriminates between correct and incorrect samples. A predictor is calibrated if predictions assigned confidence $p$ are correct with frequency $p$, and Expected Calibration Error (ECE) estimates this mismatch by binning examples by confidence and measuring the weighted gap between bin accuracy and bin confidence \citep{guo2017calibration}. Discrimination asks whether the confidence ranking separates correct from incorrect predictions, which is the property needed for selective prediction. We also report the Brier score, a proper scoring rule defined as the mean squared error between confidence and correctness.

\subsection{Confidence Driven Multiagent Debate}
Multiagent debate is a common collaborative reasoning framework in which multiple LLM agents exchange arguments before producing final answers \citep{du2024improving, Liang2024encouraging}. Debate is often motivated by the possibility that interaction can expose errors, diversify reasoning, and reduce premature commitment to an incorrect answer. Recent work suggests that the gains from multiagent debate are unstable and often come more from aggregation than from the debate's information exchange itself \citep{choi25debate, wynn2025talk, zhu2026demystifying, liu2026breaking}.

Several recent methods bring confidence into debate. Some expose each agent's confidence, allowing other agents or an aggregator to weight claims by reliability \citep{lin2025enhancing, yoffe2025debunc, bai2024confidencecal}. Others use confidence as a routing signal to decide whether debate should be triggered \citep{fan2026imad,eo2025debate}, or study whether deliberation improves the calibration of individual agents \citep{yang2024confidence, pandey2026refine}.

\section{Methods}
In this section, we formalize our multiagent confidence composition. We consider a two-agent setting with models $M_1$ and $M_2$. For a given input $x$, each agent $M_i$ ($i \in \{1,2\}$) first independently produces its zero-shot outputs at step $s=0$: a prediction $y_{i,0} \in \mathcal{Y}$, a supporting chain-of-thought $\mathrm{CoT}_{i,0}$, and an initial confidence estimate $c_{i,0} \in [0,1]$. After one round of interaction ($s=1$), each agent updates its outputs to $(y_{i,1}, \mathrm{CoT}_{i,1}, c_{i,1})$, which we refer to as the post-debate outputs.

Given the predictions $y_{i,s}$ and confidence scores $c_{i,s}$ across agents $i \in \{1,2\}$ and steps $s \in \{0,1\}$ (where $s=0$ denotes zero-shot and $s=1$ denotes post-debate), a Confidence-Aware multiagent Protocol aggregates these signals to output a final system prediction $\hat{y}_{\mathrm{P}} \in \mathcal{Y}$ and an aggregated confidence score $\hat{c}_{\mathrm{P}} \in [0,1]$.

\subsection{Confidence-Gated Fusion (CGF)} CGF is a two step setup. The first step is confidence gating, in which each agent independently decides whether to retain its zero-shot prediction or switch to its debated prediction. Then the protocol performs a probability fusion on the two surviving streams.

\paragraph{I. Confidence Gating.} In the first step, the transition to the debated output is allowed if the debate process increases that agent's confidence by more than a learned threshold $\tau_i$ in logit space: $\operatorname{logit}(c_{i,1}) - \operatorname{logit}(c_{i,0}) > \tau_i$. Upon a significant increase in confidence, the model will report $\{y_{i,1}, c_{i,1}\}$; otherwise, $\{y_{i,0}, c_{i,0}\}$.

Asymmetric-CGF (A-CGF) is another variant of this setup; rather than conducting a full symmetric debate across all cases, we initiate a one-sided debate (or persuasion step) whenever a large initial confidence imbalance exists between agents: $\lvert \operatorname{logit}(c_{1,0}) - \operatorname{logit}(c_{2,0}) \rvert > \gamma$. In this regime, the significantly more confident agent presents its rationale to convince the less confident agent, while symmetric debate is reserved for cases with initially balanced confidence. Following this interaction, A-CGF applies the exact same confidence gating.

\paragraph{II. Probability Fusion.}
After the confidence-gated updates, we apply a probability fusion step on the two surviving streams (for both CGF and A-CGF). Let $c_1$ and $c_2$ be the surviving confidences, and let
\vspace{-3mm}
\[
\ell_i = \operatorname{logit}(c_i) = \log\frac{c_i}{1-c_i}.
\]

Naive fusion treats the two streams as independent evidence. When they agree on the answer, it reports the posterior probability that the shared prediction is correct given agreement under independent evidence:
\[
c_{\text{naive}} = 
\sigma(\ell_1+\ell_2) = 
\frac{c_1c_2}{c_1c_2+(1-c_1)(1-c_2)}.
\]

When they disagree, it selects the stream with larger confidence and reports the posterior probability that Stream 1 (here $c_1>c_2$) is correct given that exactly one stream is correct:

\[
c_{\text{naive}} = \sigma\left(\lvert \ell_1-\ell_2\rvert\right) = \frac{c_1(1-c_2)}{c_1(1-c_2) + c_2(1-c_1)}
\] 

Refer to Appendix Section~\ref{sec:eq_proof} for the equivalence of the above statements.

\paragraph{Dependence Correction.}
While naive fusion treats the two surviving streams as conditionally independent, multiagent interaction inevitably introduces information redundancy and correlated errors. To account for this dependence, we introduce a shared discount parameter $w \in (0,1]$ that scales the total log-odds evidence: $c_{\text{discounted}} =$ $\sigma\left(w(\ell_1+\ell_2)\right)$ for agreement, and $c_{\text{discounted}} = \sigma\left(w\lvert \ell_1-\ell_2\rvert\right)$ for disagreement.

$w < 1$ indicates that the two agents partly repeat the same evidence, so they do not overcount their combined confidence excessively. For CGF, using the same positive $w$ for both streams also preserves ordering. If example A had more fused evidence than example B before discounting, it still has more after multiplying both by $w$. That is why we expect no change in discrimination (AUARC), while calibration (ECE and Brier) improves. Mathematically, this can be viewed as a post hoc temperature calibration, with $T = 1/w$.

\paragraph{Weighted Stream Voting (WSV).} We compare another family of confidence composition with our CGF method. WSV treats the four (agent, stream) outputs as independent pieces of evidence about the true label. Each piece contributes a learned weight $w_{i,s} \geq 0$ that controls how much the system trusts that source. For every candidate label $y_k$ that appears among the four predictions, the protocol computes a score by summing the weighted logit-confidences ($w_{i,s} \cdot \mathrm{logit}(p_{i,s})$) of all streams that predicted $y_k$. The final prediction is the candidate with the highest score, and the routed confidence is softmax-normalized to prevent inflated confidence. $w_{1,0}, w_{2,0}, w_{1,1}, w_{2,1}$ are the learned parameters. Final answer is $\hat{y} = \arg\max_{k} \text{confidence}(y_k)$
with the associated confidence $\hat{c} = \mathrm{softmax}\bigl(\text{confidences}\bigr)_{\arg\max_k}$. 

\paragraph{Debate Baselines}
We consider MAD and MAD on disagreement (MAD-D) as the baseline. In our MAD implementation, agents engage in a single-turn exchange of their reasoning. Each agent $M_i$ receives the peer's zero-shot chain-of-thought $\text{CoT}_{j,0}$ ($j \neq i$) and is prompted to produce an updated prediction $y_{i,1}$. The debated prediction $\{y_{i,1}, c_{i, 1}\}$ is used as the final answer for each model. A restriction of standard MAD limits debate to samples where the agents' initial predictions diverge ($y_{1,0} \neq y_{2,0}$). When agents initially agree, the debate round is bypassed, and $y_{i,0}$ is reported as the final answer. Similar to standard MAD, this version also produces separate responses and confidences for each model.

\subsection{Confidence Estimation}
We explore the self-report and sequence probability families of confidence estimators. Our self-report confidence estimator elicits a verbalized score by prompting the model to rate its certainty after producing its answer, and then averaging the results over three repetitions with a slightly higher model temperature (0.3) \cite{tian2023just, xiong2023verb}. The sequence probability estimator computes the mean log-likelihood over the generated token sequence, normalized by length to control for output length bias \cite{lin2023generating}, followed by a token-bias correction. Bias correction step adjusts sequence probability by subtracting the model's content-free prior $\log P_\theta(\hat{y}\mid x_{null})$ from the conditional score: $C(x,\hat{y}) = \log P_\theta(\hat{y}\mid x) - \log P_\theta(\hat{y}\mid x_{null})$, measuring how much input $x$ increases the likelihood of $\hat{y}$ beyond baseline, attempting to measure the abstract meaning of the answer rather than the tokens probability~\cite{zhao2021calibrate, holtzs}.

\paragraph{Confidence Calibration}
For all confidence-composition methods and baselines, each model's raw confidence is first passed through a beta calibration step. The calibrator is fitted to a small validation. Its role is to transform raw confidence scores into probabilities that better match empirical correctness before the protocol combines them.

\textbf{Beta calibration} is a post-hoc parametric method that applies an affine transformation to the log-odds of the raw probabilities \citep{kull2017beta}. For binary probability inputs $p$, the calibrated score is formulated as: $\tilde{c} = \sigma (a \ln p - b \ln(1 - p) + c)$ where $\sigma$ is the Sigmoid function, $a, b \ge 0$, and $c$ are scalar parameters. The transformation learns to minimize the negative log-likelihood (NLL) to predict the sample's correctness.

\subsection{Optimization}

The protocols' parameters $\theta$, such as confidence-gating thresholds, dependency discounts, and WSV weights, are learned during the optimization step. The routing step is non-differentiable; therefore, we optimize it using TPE\footnote{Tree-structured Parzen Estimator, Optuna implementation} with early stopping. The objective combines NLL as the performance term, since answer correctness remains the primary objective of a multiagent system, a Brier penalty guarding against correctness gained at the cost of calibration, and a regularizer toward neutral parameters:

\vspace{4mm}
$\hat{\theta}=\arg\min_{\theta}\;\; \underbrace{\mathrm{NLL}(\theta)}_{\text{performance}}+
\underbrace{
\lambda_{\mathrm{Brier}}\cdot \mathrm{Brier}(\theta)}_{\text{calibration}}+\\
\underbrace{\lambda_{\mathrm{anchor}}\cdot \lVert\theta - \theta_0\rVert_2^2}_{\text{regularization}}$
\vspace{4mm}

We fix $\lambda_{\mathrm{Brier}} = \lambda_{\mathrm{anchor}} = 0.1$ across all experiments. The Brier term discourages routing solutions that improve correctness but produce miscalibrated outputs, and the anchor term shrinks thresholds toward neutral defaults to prevent overfitting on the small validation split. We use separate validation data for the initial calibration and parameter optimization.

\section{Experimental Setup}
\paragraph{Models and Debater Pairs.}
We evaluate fourteen open-source models across five backbone families: LLaMA \cite{grattafiori2024llama} (3.2-3B, 3.1-8B), Mistral (Mistral-7B, Ministral-8B), Phi \cite{abdin2024phi} (3-Medium-14B, 4-Mini-3.8B, 4-14B), Gemma \cite{team2024gemma} (2-9B, 3-4B, 3-12B), and Qwen \cite{bai2023qwen} (2.5-7B, 2.5-14B, 2.5-32B, 3-4B), all instruction-tuned variants. We first evaluate each model individually on all five benchmarks and record zero-shot performance on the validation split. For each benchmark, we select homogeneous and heterogeneous pairs that differ in capability. We chose both pairs with similar zero-shot accuracy and pairs with one high and one low accuracy to ensure our results are not tied to any particular capability gap or size regime. Selected pairs and their initial performances can be found in Appendix Table~\ref{tab:debate_pairs}.

\paragraph{Datasets.}
We evaluate across five benchmarks spanning four task types to cover a variety of tasks with varying levels of ambiguity. For scientific question answering, we use \textbf{MMLU} \cite{MMLU}, a broad benchmark covering diverse academic subjects, and we chose the college-level subset. For stance detection, we use \textbf{EZStance} \cite{ezstance} and \textbf{VAST} \cite{allaway2020zero}, two datasets that differ meaningfully from factual QA in terms of ambiguity. For natural language inference–style Boolean question answering, we use \textbf{BoolQ} \cite{boolq}, with questions in a Boolean answer format. Finally, we include \textbf{HatEval} \cite{hateval}, a dataset for hate speech detection.

\paragraph{Implementation Notes.}
We apply self-consistency sampling (prompting each model three times and aggregating outputs) to obtain reliable answers and confidence estimates, and tune prompts to be minimal and uniform across models. Prompts for all of our tasks are available in the Appendix Section \ref{sec:prompts}. Our prompts are designed to simulate collaborative reasoning rather than competitive debate.

\subsection{Evaluation}
Our primary confidence-quality metric is AUARC, the area under the accuracy--rejection curve. AUARC measures whether a system ranks correct predictions above incorrect ones. It is computed by progressively rejecting the least confident predictions and measuring the accuracy of the retained predictions at each coverage level. Higher AUARC indicates a more useful confidence signal for selective prediction.

We measure calibration using expected calibration error (ECE) and Brier score. ECE partitions predictions into equal-width confidence bins and computes the bin-size-weighted absolute difference between the mean confidence and empirical accuracy in each bin \citep{guo2017calibration}. Lower ECE indicates that reported confidence more closely matches observed accuracy.

Unlike ECE, which measures the mismatch between confidence and empirical accuracy, Brier\citep{brier1950verification} is a proper scoring rule that reflects both calibration and the useful variation in confidence across correct and incorrect answers\citep{murphy1973new}. Brier score is the mean squared difference between each confidence value and its binary correctness label, so lower values indicate more accurate probabilistic predictions. 

We report both raw metrics and their temperature-calibrated versions, t-ECE and t-Brier. We fit a temperature tuning to each system's output confidences on the validation split by minimizing the negative log-likelihood, and apply it unchanged to the test split. Temperature scaling is monotone and therefore preserves confidence rankings and AUARC, allowing t-ECE and t-Brier to measure reliability after a final one-parameter calibration step~\citet{tian2023just}. 

For evaluating the zero-shot, MAD, and MAD-D models, we track the model with higher performance on the validation set, since, intuitively, a MAD pair outperforms zero-shot if the better-performing model after debate outperforms the better-performing model in zero-shot. The three introduced protocols report a single response and confidence; therefore, no selection is needed. For simplicity, we report each protocol's performance relative to the validation-selected zero-shot performance.

\paragraph{Statistical Significance.}

We assess whether a protocol consistently improves over each baseline using exact one-sided paired sign tests. Each model-pair and dataset setting provides one paired observation. The sign test is a nonparametric test designed for paired comparisons in which improvement magnitudes are not directly comparable across settings \citep{dixon1946sign}. It uses only the direction of the difference, which is appropriate here because a gain on one benchmark need not have the same numerical meaning as an equally sized gain on another benchmark.

Within each of the 30 settings, we evaluate the protocol and baseline on the same held-out test examples and compute the difference in their metrics. A setting favors the protocol when this difference is positive. The sign test then asks whether positive differences occur more often than expected by chance across the 30 settings. We run the analysis separately for sequence probability and self-reported confidence. For each metric, we test CGF and A-CGF against ZS, MAD, and MAD-D under both confidence estimators, giving 12 comparisons. We control the family-wise error rate across these comparisons using Holm's procedure \citep{holm1979}.

\section{Results}
\begin{table*}[t]
\centering
\small
\resizebox{0.90\textwidth}{!}{%
\begin{threeparttable}
\begin{tabular}{lcc|cc|cc|cc|cc}
\toprule
\textbf{Dataset:} & \multicolumn{2}{c}{\textbf{BoolQ}} & \multicolumn{2}{|c}{\textbf{VAST}} & \multicolumn{2}{|c}{\textbf{EZStance}} & \multicolumn{2}{|c}{\textbf{HatEval}} & \multicolumn{2}{|c}{\textbf{MMLU}} \\
\cmidrule(lr){2-3} \cmidrule(lr){4-5} \cmidrule(lr){6-7} \cmidrule(lr){8-9} \cmidrule(lr){10-11}
Protocol & $\Delta$AUARC$\uparrow$ & t-Brier$\downarrow$ & $\Delta$AUARC$\uparrow$ & t-Brier$\downarrow$ & $\Delta$AUARC$\uparrow$ & t-Brier$\downarrow$ & $\Delta$AUARC$\uparrow$ & t-Brier$\downarrow$ & $\Delta$AUARC$\uparrow$ & t-Brier$\downarrow$ \\
\midrule
\multicolumn{11}{l}{Sequence Probability Confidence Estimator} \\
ZS & +0.00 & 0.154 & +0.00 & 0.215 & +0.00 & 0.234 & +0.00 & 0.246 & +0.00 & 0.242 \\
MAD & -1.60 & 0.174 & -0.90 & 0.221 & -5.27 & 0.238 & +0.89 & 0.243 & +7.63 & 0.227 \\
MAD-D & +7.12 & 0.147 & +2.34 & 0.210 & +1.59 & 0.231 & \textbf{+7.01} & 0.242 & +4.41 & 0.237 \\
WSV & +10.40 & 0.142 & -0.10 & 0.217 & +2.17 & 0.235 & +6.60 & 0.253 & \textbf{+15.17} & \textbf{0.220} \\
CGF & \textbf{+10.99} & \textbf{0.132} & \textbf{+6.25} & \textbf{0.203} & \textbf{+4.55} & \textbf{0.227} & +6.27 & \textbf{0.237} & +9.35 & 0.232 \\
A-CGF & +10.79 & \textbf{0.132} & +5.59 & \textbf{0.203} & +4.23 & \textbf{0.227} & +6.05 & \textbf{0.237} & +11.22 & 0.229 \\
\midrule
\multicolumn{11}{l}{Self-Report Confidence Estimator} \\
ZS & +0.00 & 0.149 & +0.00 & 0.212 & +0.00 & 0.231 & +0.00 & 0.241 & +0.00 & 0.237 \\
MAD & -1.54 & 0.169 & -2.16 & 0.212 & -5.10 & 0.228 & -8.44 & 0.246 & +8.36 & \textbf{0.214} \\
MAD-D & +2.49 & 0.139 & +0.85 & 0.206 & +1.58 & 0.226 & +1.27 & 0.241 & +2.14 & 0.231 \\
WSV & +7.14 & 0.140 & -2.52 & 0.216 & -2.58 & 0.235 & +0.30 & 0.245 & \textbf{+12.51} & 0.219 \\
CGF & +7.59 & 0.133 & \textbf{+7.79} & \textbf{0.198} & \textbf{+2.71} & \textbf{0.224} & +1.41 & \textbf{0.234} & +3.98 & 0.223 \\
A-CGF & \textbf{+7.78} & \textbf{0.131} & +6.27 & 0.201 & +2.58 & 0.225 & \textbf{+1.52} & 0.235 & +5.63 & 0.222 \\
\bottomrule
\end{tabular}
\caption{\textbf{System-level confidence quality across five benchmarks.} AUARC is the percentage-point change relative to the validation-selected best single zero-shot agent (higher is better). t-Brier is the absolute Brier score after every system, including the baselines, receives the same one-parameter temperature calibration fitted on validation data (lower is better). Results are averaged over six model pairs per dataset. Bold marks the best value within each dataset and confidence-estimator column.}
\vspace{-3mm}
\label{tab:results_main}
\end{threeparttable}
}
\end{table*}

\begin{figure*}[t]
    \centering
    \includegraphics[width=0.80\textwidth]{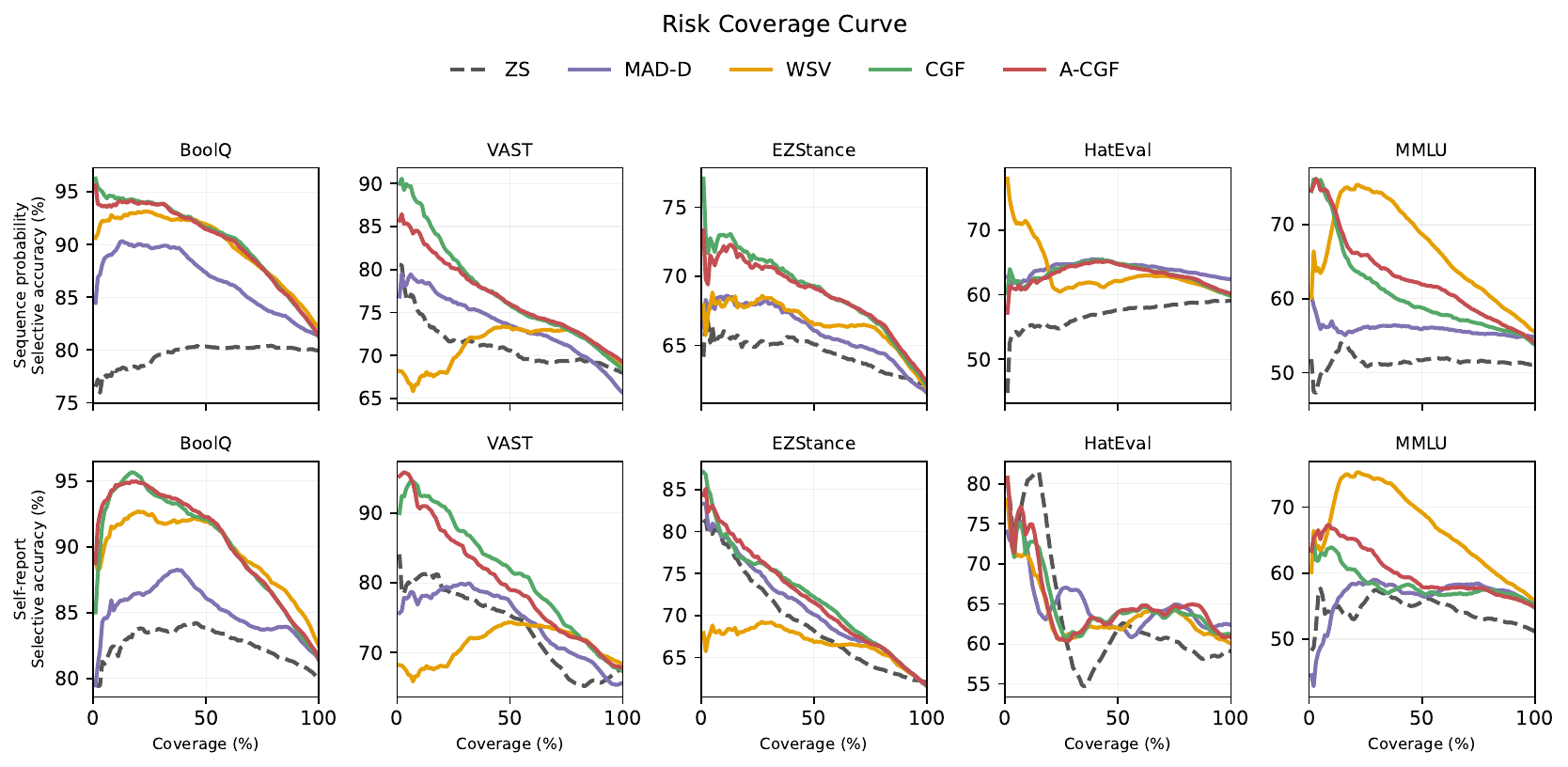}
    \caption{\textbf{Risk--coverage curves.} For every model pair, we rank test predictions by confidence and compute accuracy after retaining only the most confident fraction of answers at each coverage level; we then average these curves across pairs. A higher curve means that, at the same coverage, the system retains more correct answers and defers more likely mistakes.}
    \vspace{-5mm}
    \label{fig:risk_coverage}
\end{figure*}

\paragraph{Confidence composition yields more discriminative system-level confidence.} Table~\ref{tab:results_main} shows that CGF and A-CGF consistently improve the ranking quality of the final system's confidence with both confidence estimators. Across all 30 model-pair settings, both protocols significantly outperform ZS, MAD, and MAD-D on AUARC under both confidence estimators (with Holm-adjusted paired sign-test $p<.01$ values; Appendix Table~\ref{tab:significance}); the median AUARC gains over MAD-D are \(+2.42\) and \(+2.16\) points for CGF, and \(+2.40\) and \(+2.23\) points for A-CGF, under sequence probability and self-report confidence, respectively. CGF and A-CGF attain the largest AUARC gain in seven of the ten dataset and estimator conditions (Table~\ref{tab:results_main}). Standard MAD is substantially less stable and reduces AUARC on BoolQ, VAST, and EZStance under sequence probability, and on all but MMLU under self-report confidence. MAD-D recovers some of this discrimination, but is generally surpassed by confidence composition. The exceptions are informative. On sequence-probability HatEval, MAD-D has the largest mean AUARC gain, while on MMLU, where debate itself is beneficial, WSV has the largest gain under both estimators. The descriptive Friedman analysis likewise places CGF and A-CGF first or second in average rank for AUARC under both estimators (Appendix Table~\ref{tab:main_rankings}). Appendix Figures~\ref{fig:auarc_ladder} and~\ref{fig:tbrier_ladder} show the corresponding per-model-pair results.

Figure~\ref{fig:risk_coverage} shows risk--coverage curves for the final system confidences. CGF and A-CGF generally lie above the single-agent ZS and selective-debate MAD-D curves, showing that their confidence signals are better at ranking correct answers above errors. MAD-D is competitive on HatEval, and WSV is particularly strong on MMLU, consistent with the dataset-level AUARC results in Table~\ref{tab:results_main}.

\paragraph{Composed confidences are better probability estimates.} Table~\ref{tab:results_main} shows that CGF and A-CGF achieve the lowest or tied-lowest t-Brier in eight of the ten datasets and estimator conditions. Across the 12 protocol--baseline comparisons, CGF or A-CGF achieves significantly lower t-Brier than ZS, MAD, or MAD-D in 11 cases (paired sign tests with Holm-adjusted $p<.05$; Appendix Table~\ref{tab:significance}). Thus, the AUARC advantage is not limited to confidence ranking. The composed signals also yield stronger final probability estimates under a proper scoring rule. Raw ECE and Brier results before temperature scaling are reported separately in Appendix Table~\ref{tab:dependence_reliability_by_dataset}.

\paragraph{Task accuracy remains competitive.} 

We compared the accuracy of our protocols against the standard debate and recent state-of-the-art confidence-based debate methods (\cite{yoffe2025debunc, lin2025enhancing}) to ensure that log-odds fusion decision-making does not degrade the system's overall performance. (Table~\ref{tab:f1_safeguard}) Standard debate is strongly task-dependent. MAD reduces F1 on BoolQ, VAST, EZStance, and HatEval by 3.25--8.20 points, while improving MMLU by 4.37 points. Restricting debate to initial disagreements reduces some of these losses and improves HatEval and MMLU, but remains below zero-shot on BoolQ, VAST, and EZStance. These patterns are consistent with prior evidence that debate does not provide uniform accuracy gains across tasks \citep{choi25debate,wynn2025talk,wu2025can}; many prior evaluations emphasize factual or scientific question answering, where a single answer is more clearly verifiable, whereas BoolQ and stance detection involve greater ambiguity and label interpretation.

\begin{table}[h]
\centering
\small
\resizebox{0.475\textwidth}{!}{%
\begin{threeparttable}
\begin{tabular}{l|ccccc}
\toprule
Protocol & \textbf{BoolQ} & \textbf{VAST} & \textbf{EZStance} & \textbf{HatEval} & \textbf{MMLU} \\
\midrule
\multicolumn{6}{l}{Confidence Independent} \\
ZS & +0.00 & +0.00 & \textbf{+0.00} & +0.00 & +0.00 \\
MAD & -4.82 & -6.02 & -8.20 & -3.25 & \textbf{+4.37} \\
MAD-D & -0.16 & -2.21 & -0.64 & \textbf{+4.81} & +3.14 \\
\midrule
\multicolumn{6}{l}{Sequence Probability Confidence Estimator} \\
ConfMAD-final$^\dagger$ & -5.05 & -6.07 & -7.92 & -3.41 & +4.10 \\
DebUnc-2agent-vote$^\dagger$ & -20.29 & -11.71 & -10.88 & -3.97 & +2.43 \\

WSV & \textbf{+0.74} & +0.74 & -0.58 & +1.81 & +3.57 \\
CGF & -0.12 & +0.38 & -0.30 & +1.59 & +1.93 \\
A-CGF & -0.16 & \textbf{+1.42} & -0.01 & +1.96 & +2.03 \\
\midrule
\multicolumn{6}{l}{Self-Report Confidence Estimator} \\
ConfMAD-final$^\dagger$ & -5.31 & -6.16 & -7.25 & -2.77 & \textbf{+4.41} \\
DebUnc-2agent-vote$^\dagger$ & -20.29 & -11.71 & -10.88 & -3.97 & +2.43 \\
WSV & \textbf{+1.28} & \textbf{+0.42} & -0.94 & +1.99 & +3.81 \\
CGF & +0.50 & -0.57 & -0.92 & +3.40 & +3.44 \\
A-CGF & +0.56 & -0.06 & -0.88 & +2.37 & +3.09 \\
\bottomrule
\end{tabular}
\caption{\textbf{Task accuracy across five benchmarks.} Values are mean weighted-F1 changes in percentage points relative to the validation-selected best single zero-shot agent, averaged over 6 pairs per dataset. Higher is better; bold marks the best value within each dataset and confidence-estimator block. $^\dagger$ConfMAD-final and DebUnc-vote are two confidence-based debate baselines in the literature. Implementation details of the debate baselines are in Section~\ref{sec:f1baseline}.}
\vspace{-3mm}
\label{tab:f1_safeguard}
\end{threeparttable}

}
\end{table}

The confidence-composition protocols are not uniformly more accurate than the strongest debate baseline. Instead, they generally avoid the large F1 losses of unrestricted MAD while providing the stronger system-level confidence signal shown in Table~\ref{tab:results_main}. They yield positive mean changes in several conditions on BoolQ and VAST, improve HatEval and MMLU under both confidence estimators, and remain near the zero-shot reference on EZStance, where the gains are small or negative. MAD-D remains the strongest method on HatEval, whereas MAD remains strongest on MMLU. Thus, we interpret F1 as an accuracy safeguard rather than as evidence that confidence composition universally improves the final answer.

\begin{figure*}[t]
    \centering
    \includegraphics[width=0.8\textwidth]{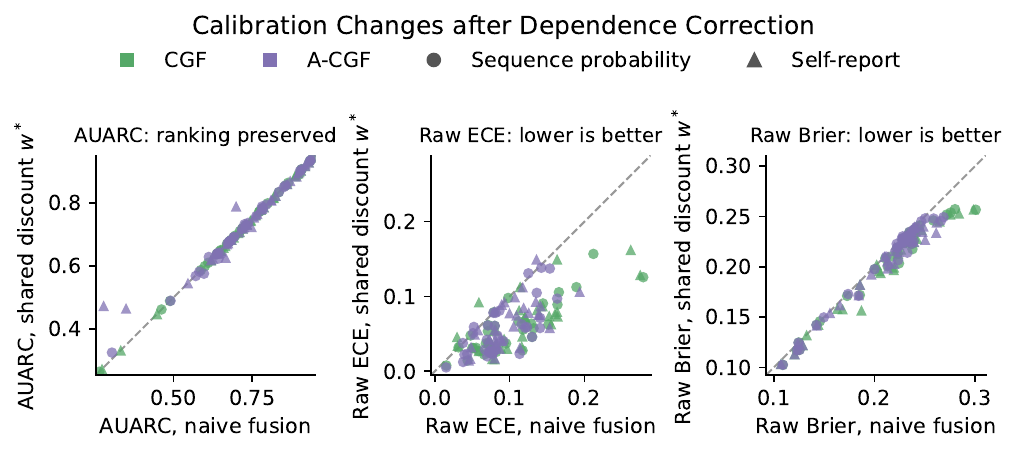}
    \caption{\textbf{Dependence correction.} Each point is a model-pair/protocol/estimator setting; the dashed line denotes no change. CGF lies on the AUARC identity line because a positive shared log-odds discount is monotone, while raw Brier generally falls below the identity line. A-CGF may deviate because one branch reports an unfused confidence.}
    \label{fig:e1_invariance}
    \vspace{-5mm}
\end{figure*}

\paragraph{Calibration and Correction for Dependence Assumption}

Table~\ref{tab:dependence_reliability_by_dataset} shows that CGF and A-CGF have higher raw ECE than MAD and MAD-D, even though all input streams are beta-calibrated. The reason is that MAD and MAD-D simply select one calibrated stream, while WSV, CGF, and A-CGF transform calibrated inputs through aggregation or fusion. After one-parameter output temperature scaling, t-ECE and t-Brier return to baseline-level or better values in most settings.

Naive log-odds fusion treats the surviving confidence streams as conditionally independent, even though interactions make them correlated. We correct this scale error by fitting a shared discount factor $w$ on validation data and applying it to the pooled log-odds evidence. Across 30 model pairs, two confidence estimators, and two fused protocols, the fitted value was below 1 in 118 of 120 settings, with a median $w^*$ of 0.65. Thus, the validation data consistently prefer counting the fused evidence at roughly two-thirds of the strength implied by independence.


The correction improves calibration without changing the final decisions. For CGF, it also preserves the confidence ranking almost exactly, with maximum absolute AUARC change of 0.00148, while raw ECE drops from 0.100 to 0.055 under sequence probability and from 0.112 to 0.063 under self-report. Raw Brier improves as well, and Figure~\ref{fig:e1_invariance} shows the same pattern visually: CGF remains on the AUARC identity line, while raw ECE and Brier move below it after discounting. Because this discount is equivalent to an output temperature with $T=1/w$, t-ECE and t-Brier change little after discounting; the main cost of the violated independence assumption is a one-scalar confidence-scale error.

\paragraph{Ablation study.} 
We compared CGF with two ablation versions and two voting variants to understand the source of improvements. For CGF ablation variants, we (1) applied the log-odds combination on zero-shot streams without debate, and (2) applied confidence gating on one round debate and chose the response with higher confidence. For voting variants, we used (1) weighted voting between the zero-shot streams and (2) weighted voting between zero-shot and debate (same as WSV).

The Holm sign-test (Appendix Table~\ref{tab:e2_tests}) showed that gating and fusion are complementary, and gate-only or Fuse on zero-shot are substantially less effective when used solely in both AUARC and t-Brier. The one-sided sign test did not show evidence of a performance difference between CGF and WSV; fusion on zero-shot streams is significantly less effective.

\section{Conclusion}

Multiagent, multi-round language model systems produce intermediate responses and confidence signals at different stages of the process. Using a two-agent debate as a controlled environment, we studied how these correlated signals can be combined into a single confidence estimate for the final answer. We evaluated two families of solutions: weighted voting over agent outputs across rounds of debate (WSV), and log-odds probability fusion applied after each model revises its belief under a confidence condition (CGF and A-CGF). 

We provided evidence that reporting the confidence of the debate's winning stream is miscalibrated and substantially less discriminative between correct and incorrect samples because it ignores evidence from the other model and the earlier debate round. Our confidence composition methods showed significantly higher discrimination (AUARC) and better probability quality (t-Brier). WSV is strongest on the least ambiguous task but less consistent across settings. These gains do not come at the cost of task accuracy; our protocols avoid the large F1 losses that unrestricted debate incurs on the more ambiguous labeling tasks.

Although the log-odds independence assumption can cause miscalibration by double-counting evidence in a structurally dependent setting such as debate, a discount coefficient trained in log space was sufficient to mitigate this effect. Additionally, we showed that the improvements from confidence gating combined with fusion are not achievable with either mechanism alone. These results separate two roles that are often conflated. Composing evidence across streams determines how well a system's confidence discriminates, while dependence correction determines how well it is calibrated.

\section{Limitations} 
This work is limited to the two-agent, one-round setting, as it is the most controlled setting for studying confidence composition.
Two-agent debate is the common minimal form of multiagent interaction and has a relatively low inference cost.
We therefore analyze this setting in depth, using 30 model pairs and tasks with varying degrees of ambiguity. Additionally, the main components of our approach can extend beyond this setup. Confidence gating can be applied at later rounds by deciding whether each new response should replace the previous one, and the log-odds fusion rule can extend to more streams (e.g. $\sigma(\operatorname{logit}(c_1)+\operatorname{logit}(c_2)+\operatorname{logit}(c_3))$). However, moving to three or more agents also introduces many additional design choices in the debate topology, such as whether all agents interact symmetrically, whether one agent acts as a judge or sub-agent, and how information flows across rounds. These choices substantially expand the experimental space, so we leave a systematic study of larger and multi-round systems to future work.

\bibliography{custom}

\appendix

\counterwithin{figure}{section}
\counterwithin{table}{section}

\FloatBarrier
\clearpage
\section{Reproducibility Details}
\label{app:reproducibility}
\begin{table*}[b!]
\centering
\resizebox{\textwidth}{!}{%
\begin{tabular}{c|c c|c c|c c|c c}
\toprule
\textbf{Task -- Dataset} & \textbf{Model 1} & \textbf{Model 2} &
\textbf{F1$_{M1}$} & \textbf{F1$_{M2}$} &
\textbf{AUARC$_{verb,M1}$} & \textbf{AUARC$_{verb,M2}$} &
\textbf{AUARC$_{seq,M1}$} & \textbf{AUARC$_{seq,M2}$} \\
\midrule

\multirow{6}{*}{Scientific QA -- MMLU}
  & Gemma 2-9B & Gemma 3-12B & 0.6044 & 0.6318 & 0.6618 & 0.6444 & 0.6159 & 0.6375 \\
  & Gemma 2-9B & Qwen 2.5-7B & 0.6044 & 0.6282 & 0.6618 & 0.6421 & 0.6159 & 0.7374 \\
  & LLaMA 3.1-8B & Gemma 3-12B & 0.5118 & 0.6307 & 0.5213 & 0.6402 & 0.5249 & 0.6355 \\
  & LLaMA 3.2-3B & Mistral 7B & 0.4267 & 0.4274 & 0.4509 & 0.4090 & 0.4500 & 0.4881 \\
  & Mistral 7B & Ministral 8B & 0.4274 & 0.4381 & 0.4090 & 0.3791 & 0.4881 & 0.2849 \\
  & Phi 4-Mini & Qwen 3-4B & 0.5205 & 0.6193 & 0.4507 & 0.5582 & 0.6402 & 0.6217 \\
  \midrule

  \multirow{6}{*}{Stance Detection -- EZStance}
  & Gemma 2-9B & Gemma 3-12B & 0.6026 & 0.5747 & 0.6826 & 0.6971 & 0.6490 & 0.6489 \\
  & LLaMA 3.1-8B & Phi 3-Medium & 0.6178 & 0.5747 & 0.6998 & 0.6907 & 0.6253 & 0.6480 \\
  & LLaMA 3.2-3B & Qwen 2.5-14B & 0.5619 & 0.6231 & 0.6615 & 0.7281 & 0.6058 & 0.6464 \\
  & Mistral 7B & Gemma 2-9B & 0.6102 & 0.6026 & 0.6661 & 0.6826 & 0.6283 & 0.6490 \\
  & Ministral 8B & Qwen 2.5-32B & 0.6364 & 0.6421 & 0.7034 & 0.7485 & 0.5950 & 0.6774 \\
  & Phi 4-Mini & Gemma 3-4B & 0.5202 & 0.5506 & 0.6546 & 0.6323 & 0.6634 & 0.6375 \\
  \midrule

  \multirow{6}{*}{Binary QA -- BoolQ}
  & Gemma 2-9B & Gemma 3-4B & 0.7657 & 0.8223 & 0.7307 & 0.8140 & 0.4688 & 0.8281 \\
  & LLaMA 3.1-8B & Gemma 2-9B & 0.7448 & 0.7657 & 0.8331 & 0.7307 & 0.8242 & 0.4688 \\
  & LLaMA 3.1-8B & LLaMA 3.2-3B & 0.7448 & 0.6813 & 0.8331 & 0.7172 & 0.8242 & 0.6463 \\
  & Mistral 7B & Qwen 2.5-7B & 0.8395 & 0.8421 & 0.8412 & 0.9127 & 0.8399 & 0.8885 \\
  & Ministral 8B & Phi 3-Medium & 0.8345 & 0.8771 & 0.8254 & 0.8966 & 0.8900 & 0.9074 \\
  & Phi 4-Mini & Gemma 3-12B & 0.8139 & 0.8498 & 0.8484 & 0.8527 & 0.9144 & 0.8718 \\
  \midrule

  \multirow{6}{*}{Hate Speech Detection -- HatEval}
  & Gemma 3-12B & Qwen 2.5-32B & 0.6826 & 0.6170 & 0.6796 & 0.6736 & 0.6967 & 0.5456 \\
  & LLaMA 3.1-8B & Gemma 3-4B & 0.5875 & 0.5509 & 0.5907 & 0.6366 & 0.5673 & 0.5398 \\
  & LLaMA 3.1-8B & LLaMA 3.2-3B & 0.5875 & 0.6013 & 0.5907 & 0.5794 & 0.5673 & 0.5990 \\
  & LLaMA 3.2-3B & Qwen 2.5-7B & 0.6013 & 0.5969 & 0.5794 & 0.6080 & 0.5990 & 0.6199 \\
  & Mistral 7B & Gemma 2-9B & 0.6207 & 0.5038 & 0.5799 & 0.6498 & 0.6858 & 0.5179 \\
  & Phi 3-Medium & Phi 4-Mini & 0.6125 & 0.5733 & 0.5650 & 0.6424 & 0.4984 & 0.6084 \\
  \midrule

  \multirow{6}{*}{Stance Detection -- VAST}
  & LLaMA 3.2-3B & LLaMA 3.1-8B & 0.5921 & 0.5682 & 0.6562 & 0.6470 & 0.6507 & 0.5385 \\
  & Mistral 7B & Gemma 3-12B & 0.6882 & 0.6676 & 0.7239 & 0.7576 & 0.7154 & 0.7592 \\
  & Mistral 7B & Ministral 8B & 0.6882 & 0.6666 & 0.7239 & 0.7287 & 0.7154 & 0.6146 \\
  & Phi 3-Medium & Gemma 3-4B & 0.7472 & 0.5251 & 0.8622 & 0.6588 & 0.7253 & 0.5265 \\
  & Phi 4-Mini & LLaMA 3.2-3B & 0.5714 & 0.5921 & 0.7638 & 0.6562 & 0.6576 & 0.6507 \\
  & Qwen 2.5-14B & Qwen 2.5-32B & 0.7301 & 0.7556 & 0.7675 & 0.7956 & 0.7761 & 0.8089 \\

\bottomrule
\end{tabular}
}
\caption{Zero-shot performance comparison between debating model pairs across tasks and datasets. F1 and AUARC denote task performance and confidence signal discrimination, respectively. AUARC is reported using self-report and sequence probability confidence estimation methods. All models are instruction versions.}
\label{tab:debate_pairs}
\end{table*}
\vspace{-2mm}
\subsection{Computational Budget and Infrastructure}
\label{app:compute}
All of our experiments are inference-only; we do not perform any model training or fine-tuning. The only learning step is the lightweight optimization of per-stream calibrators and routing thresholds, which runs on CPU and adds negligible cost. Model inference was conducted on a single NVIDIA H200 NVL (141GB) GPU. The full set of experiments (fourteen models across five benchmarks, six debater pairs per benchmark, with self-consistency sampling of three generations per query) can be reproduced on a single NVIDIA H200 GPU in 15 days. Parameter counts for all models follow their officially released sizes (3B--32B), listed in the Experimental setup section.

\begin{table}[H]
\centering
\small
\begin{tabular}{llrr}
\toprule
\textbf{Dataset} & \textbf{Task} & \textbf{Dev} & \textbf{Test} \\
\midrule
BoolQ    & Boolean QA       &  {981}   & {2,289} \\
VAST     & Stance detection  & {1973} & {2830} \\
EZStance & Stance detection  & {2354} & {2663} \\
HatEval  & Hate speech det. & {1,000} & {2971} \\
MMLU-college     & Scientific QA & {195} & {1,789} \\
\bottomrule
\end{tabular}
\caption{Dataset statistics and split sizes used in our experiments.}
\label{tab:dataset_stats}
\end{table}

\subsection{Datasets and Statistics}
\label{app:datasets}
We evaluate on five publicly available English benchmarks spanning four task types. Table~\ref{tab:dataset_stats} reports the task and split sizes used in our experiments. For every dataset, we use a held-out validation split to fit calibrators and learn routing thresholds, and report all results on the test split; the validation split is never used for evaluation. The training set is never used in general. BoolQ and HatEval do not have a validation set; we used a random sample of 1000 from the test set as validation (random state=42). Additionally, MMLU was used with its college samples.

All datasets are standard research benchmarks used in accordance with their intended research use.

The code repository will be public upon publication.

\subsection{Initial Zero-shot Results}
\label{app:zs}

Table~\ref{tab:debate_pairs} lists the model pairs along with their initial performance and initial AUARC under both sequence probability and self-report confidence estimators.

\FloatBarrier
\clearpage
\section{Statistical Significance Test Results}
\label{app:significance}

\noindent
\begin{minipage}{\textwidth}
\centering
\small
\begin{threeparttable}

\begin{tabular}{lllrrcrrc}
\toprule
& & & \multicolumn{3}{c}{AUARC} & \multicolumn{3}{c}{t-Brier} \\
Estimator & Protocol & vs. & Median $\Delta$ & Wins & $p_{\mathrm{Holm}}$ & Median $\Delta$ & Wins & $p_{\mathrm{Holm}}$ \\
\midrule
Sequence probability & CGF & ZS & +6.11 & 30/30 & $<.0001^{**}$ & +0.85 & 29/30 & $<.0001^{**}$ \\
Sequence probability & CGF & ZS & +6.11 & 30/30 & $<.0001^{**}$ & +0.85 & 29/30 & $<.0001^{**}$ \\
Sequence probability & CGF & MAD & +8.40 & 27/30 & $<.0001^{**}$ & +1.18 & 27/30 & $<.0001^{**}$ \\
Sequence probability & CGF & MAD-D & +2.42 & 28/30 & $<.0001^{**}$ & +0.65 & 25/30 & $0.0011^{**}$ \\
Sequence probability & A-CGF & ZS & +5.96 & 30/30 & $<.0001^{**}$ & +0.89 & 30/30 & $<.0001^{**}$ \\
Sequence probability & A-CGF & MAD & +8.50 & 28/30 & $<.0001^{**}$ & +1.09 & 26/30 & $0.0002^{**}$ \\
Sequence probability & A-CGF & MAD-D & +2.40 & 28/30 & $<.0001^{**}$ & +0.73 & 27/30 & $<.0001^{**}$ \\
Self-report & CGF & ZS & +4.67 & 26/30 & $0.0002^{**}$ & +0.84 & 30/30 & $<.0001^{**}$ \\
Self-report & CGF & MAD & +5.01 & 28/30 & $<.0001^{**}$ & +0.67 & 24/30 & $0.0043^{**}$ \\
Self-report & CGF & MAD-D & +2.16 & 27/30 & $<.0001^{**}$ & +0.39 & 25/30 & $0.0011^{**}$ \\
Self-report & A-CGF & ZS & +4.13 & 28/30 & $<.0001^{**}$ & +0.79 & 30/30 & $<.0001^{**}$ \\
Self-report & A-CGF & MAD & +5.09 & 26/30 & $0.0002^{**}$ & +0.59 & 21/30 & $0.0643$ \\
Self-report & A-CGF & MAD-D & +2.23 & 27/30 & $<.0001^{**}$ & +0.29 & 25/30 & $0.0011^{**}$ \\

\bottomrule
\end{tabular}
\begin{tablenotes}[flushleft]\footnotesize
\item $^{*}p_{\mathrm{Holm}}<.05$; $^{**}p_{\mathrm{Holm}}<.01$.
\end{tablenotes}
\end{threeparttable}
\captionof{table}{\textbf{Holm's Significance test.} Each row compares a protocol with one named baseline over the same 30 model pairs; positive $\Delta$ favors the protocol. We use exact one-sided sign tests and correct the 12 comparisons within AUARC and t-Brier separately using Holm's method. AUARC is the primary endpoint; t-Brier is secondary. Sequence probability and self-report are analyzed separately because they are derived from the same generations and are not independent observations.}
\label{tab:significance}
\end{minipage}

\vspace{5mm}
\noindent
\begin{minipage}{\textwidth}
\centering
\small
\begin{threeparttable}
\begin{tabular}{llrrcp{0.40\textwidth}}
\toprule
Estimator & Metric & Blocks & Friedman $p$ & CD & Average rank (best $\to$ worst) \\
\midrule
Sequence probability & AUARC & 30 & $<.0001$ & 1.377 & A-CGF (2.28) $\;\to\;$ CGF (2.34) $\;\to\;$ WSV (2.68) $\;\to\;$ MAD-D (3.91) $\;\to\;$ MAD (4.59) $\;\to\;$ ZS (5.21) \\
\midrule
Sequence probability & t-Brier & 30 & $<.0001$ & 1.377 & A-CGF (2.34) $\;\to\;$ CGF (2.37) $\;\to\;$ MAD-D (3.50) $\;\to\;$ WSV (3.91) $\;\to\;$ MAD (4.41) $\;\to\;$ ZS (4.47) \\
\midrule
Self-report & AUARC & 30 & $<.0001$ & 1.377 & A-CGF (2.32) $\;\to\;$ CGF (2.38) $\;\to\;$ WSV (3.68) $\;\to\;$ MAD-D (3.76) $\;\to\;$ MAD (4.35) $\;\to\;$ ZS (4.50) \\
\midrule
Self-report & t-Brier & 30 & $<.0001$ & 1.377 & CGF (2.38) $\;\to\;$ A-CGF (2.47) $\;\to\;$ MAD-D (3.41) $\;\to\;$ MAD (3.74) $\;\to\;$ WSV (4.29) $\;\to\;$ ZS (4.71) \\
\bottomrule
\end{tabular}
\end{threeparttable}
\captionof{table}{\textbf{Friedman test with Nemenyi critical differences for descriptive overall rankings.} Friedman ranks the six headline systems within each of 30 model-pair blocks, separately per estimator; lower average rank is better. The Nemenyi critical difference (CD) is the minimum rank gap required for a pairwise distinction at $\alpha=.05$.}
\label{tab:main_rankings}
\end{minipage}
\vspace{5mm}

\paragraph{Friedman's Rank Test.}
In addition to Holm's sign test explained in experimental setup, to summarize the relative performance setup of all systems simultaneously, we use Friedman's test followed by Nemenyi critical-difference comparisons \citep{friedman1937ranks,nemenyi1963distribution,demsar2006statistical}. For each model-pair and dataset setting, we rank the six headline systems according to the metric of interest, with rank 1 assigned to the best system. Friedman’s test then assesses whether the average ranks differ more than expected by chance across the matched settings. When this test is significant, the Nemenyi critical difference indicates how far apart two average ranks must be before they can be distinguished at the chosen significance level. this test is reported in Table~\ref{tab:main_rankings}.

\FloatBarrier
\clearpage
\section{Calibration and Dependence Correction}

\noindent
\begin{minipage}{\textwidth}
\centering
\scriptsize
\begin{threeparttable}

\resizebox{\textwidth}{!}{%
\begin{tabular}{lcccc|cccc|cccc|cccc|cccc}
\toprule
System & \multicolumn{4}{c}{BoolQ} & \multicolumn{4}{c}{VAST} & \multicolumn{4}{c}{EZStance} & \multicolumn{4}{c}{HatEval} & \multicolumn{4}{c}{MMLU} \\
 & ECE & t-ECE & Brier & t-Brier & ECE & t-ECE & Brier & t-Brier & ECE & t-ECE & Brier & t-Brier & ECE & t-ECE & Brier & t-Brier & ECE & t-ECE & Brier & t-Brier \\
\midrule
\multicolumn{21}{l}{\textit{Sequence probability}} \\
ZS & 0.404 & 0.303 & 0.357 & 0.254 & 0.167 & 0.105 & 0.249 & 0.229 & 0.203 & 0.096 & 0.297 & 0.245 & 0.251 & 0.094 & 0.337 & 0.251 & 0.276 & 0.095 & 0.350 & 0.250 \\

MAD & 0.012 & 0.012 & 0.174 & 0.174 & 0.047 & 0.047 & 0.221 & 0.221 & 0.018 & 0.018 & 0.238 & 0.238 & 0.033 & 0.033 & 0.243 & 0.243 & 0.035 & 0.035 & 0.227 & 0.227 \\
MAD-D & 0.054 & 0.051 & 0.147 & 0.147 & 0.048 & 0.062 & 0.209 & 0.210 & 0.039 & 0.028 & 0.231 & 0.231 & 0.080 & 0.090 & 0.240 & 0.242 & 0.087 & 0.070 & 0.237 & 0.237 \\
WSV & 0.089 & 0.081 & 0.140 & 0.142 & 0.221 & 0.088 & 0.260 & 0.217 & 0.231 & 0.046 & 0.295 & 0.235 & 0.188 & 0.102 & 0.289 & 0.253 & 0.246 & 0.092 & 0.274 & 0.220 \\

\addlinespace[2pt]
CGF, naive & 0.068 & 0.039 & 0.135 & 0.132 & 0.127 & 0.064 & 0.217 & 0.203 & 0.067 & 0.028 & 0.231 & 0.227 & 0.086 & 0.051 & 0.244 & 0.237 & 0.154 & 0.095 & 0.248 & 0.232 \\
CGF, shared discount & 0.041 & 0.039 & 0.131 & 0.132 & 0.060 & 0.064 & 0.203 & 0.203 & 0.028 & 0.028 & 0.227 & 0.227 & 0.051 & 0.051 & 0.237 & 0.237 & 0.092 & 0.095 & 0.232 & 0.232 \\
A-CGF, naive & 0.065 & 0.035 & 0.135 & 0.132 & 0.118 & 0.055 & 0.215 & 0.203 & 0.065 & 0.025 & 0.231 & 0.227 & 0.091 & 0.049 & 0.245 & 0.237 & 0.109 & 0.093 & 0.232 & 0.229 \\
A-CGF, shared discount & 0.037 & 0.035 & 0.132 & 0.132 & 0.055 & 0.056 & 0.203 & 0.203 & 0.025 & 0.023 & 0.227 & 0.227 & 0.054 & 0.050 & 0.238 & 0.238 & 0.099 & 0.100 & 0.229 & 0.229 \\
\addlinespace[6pt]
\multicolumn{21}{l}{\textit{Self-report}} \\
ZS & 0.172 & 0.052 & 0.184 & 0.158 & 0.295 & 0.141 & 0.327 & 0.246 & 0.242 & 0.048 & 0.293 & 0.233 & 0.329 & 0.051 & 0.350 & 0.243 & 0.415 & 0.101 & 0.418 & 0.255 \\
MAD & 0.012 & 0.012 & 0.169 & 0.169 & 0.044 & 0.044 & 0.212 & 0.212 & 0.017 & 0.017 & 0.228 & 0.228 & 0.047 & 0.042 & 0.246 & 0.246 & 0.050 & 0.050 & 0.214 & 0.214 \\
MAD-D & 0.077 & 0.059 & 0.139 & 0.139 & 0.040 & 0.060 & 0.205 & 0.206 & 0.041 & 0.037 & 0.226 & 0.226 & 0.087 & 0.093 & 0.240 & 0.241 & 0.099 & 0.076 & 0.231 & 0.231 \\
WSV & 0.106 & 0.092 & 0.140 & 0.140 & 0.233 & 0.085 & 0.263 & 0.216 & 0.230 & 0.048 & 0.295 & 0.235 & 0.201 & 0.081 & 0.288 & 0.245 & 0.238 & 0.100 & 0.266 & 0.219 \\
\addlinespace[2pt]
CGF, naive & 0.085 & 0.055 & 0.138 & 0.133 & 0.135 & 0.075 & 0.213 & 0.198 & 0.087 & 0.039 & 0.230 & 0.224 & 0.088 & 0.046 & 0.241 & 0.234 & 0.170 & 0.095 & 0.242 & 0.223 \\
CGF, shared discount & 0.055 & 0.055 & 0.133 & 0.133 & 0.071 & 0.075 & 0.197 & 0.198 & 0.042 & 0.039 & 0.224 & 0.224 & 0.043 & 0.046 & 0.234 & 0.234 & 0.103 & 0.095 & 0.224 & 0.223 \\
A-CGF, naive & 0.081 & 0.052 & 0.135 & 0.131 & 0.123 & 0.082 & 0.212 & 0.201 & 0.071 & 0.038 & 0.229 & 0.225 & 0.098 & 0.049 & 0.242 & 0.235 & 0.118 & 0.099 & 0.227 & 0.222 \\
A-CGF, shared discount & 0.052 & 0.053 & 0.132 & 0.131 & 0.082 & 0.083 & 0.201 & 0.201 & 0.035 & 0.038 & 0.225 & 0.225 & 0.051 & 0.051 & 0.236 & 0.236 & 0.090 & 0.084 & 0.218 & 0.216 \\
\bottomrule
\end{tabular}}

\caption{\textbf{Calibration before and after dependence correction.} Mean test ECE, t-ECE, Brier, and t-Brier across model pairs within each dataset. Lower is better. The first ZS row uses raw input confidence from the identity run; its $t$ column applies output temperature scaling. Every other row uses beta-calibrated input streams. For CGF and A-CGF, the naive row uses independent fusion ($w=1$) and the shared-discount row uses $w^*$ fitted on validation data. The $t$ metrics apply a separately validation-fitted output temperature to each row.}
\label{tab:dependence_reliability_by_dataset}
\end{threeparttable}
\end{minipage}

\FloatBarrier
\section{Ablation}
\begin{minipage}{\textwidth}
\centering\small
\begin{threeparttable}

\begin{tabular}{llrccrcc}\toprule
& & \multicolumn{3}{c}{AUARC} & \multicolumn{3}{c}{t-Brier} \\
Estimator & CGF vs. & Median $\Delta$ (pp) & Wins & $p_{\mathrm{Holm}}$ & Median $\Delta$ (pp) & Wins & $p_{\mathrm{Holm}}$ \\
\midrule
Sequence probability & ZS-vote2 & +4.64 & 30/30 & $<.0001^{**}$ & +0.83 & 30/30 & $<.0001^{**}$ \\
Sequence probability & WSV & +2.61 & 24/30 & $0.0043^{**}$ & +0.49 & 24/30 & $0.0043^{**}$ \\
Sequence probability & Gate-only & +4.11 & 29/30 & $<.0001^{**}$ & +0.64 & 30/30 & $<.0001^{**}$ \\
Sequence probability & Fuse-ZS & +0.08 & 21/30 & $0.0643$ & +0.00 & 20/30 & $0.0987$ \\
\midrule
Self-report & ZS-vote2 & +2.66 & 28/30 & $<.0001^{**}$ & +0.65 & 29/30 & $<.0001^{**}$ \\
Self-report & WSV & -0.01 & 17/30 & $0.2923$ & -0.05 & 16/30 & $0.4278$ \\
Self-report & Gate-only & +2.07 & 28/30 & $<.0001^{**}$ & +0.50 & 29/30 & $<.0001^{**}$ \\
Self-report & Fuse-ZS & +0.45 & 24/30 & $0.0043^{**}$ & +0.05 & 26/30 & $0.0002^{**}$ \\
\bottomrule\end{tabular}
\begin{tablenotes}[flushleft]\footnotesize
\item $^{*}p_{\mathrm{Holm}}<.05$; $^{**}p_{\mathrm{Holm}}<.01$.
\end{tablenotes}
\end{threeparttable}
\captionof{table}{\textbf{CGF ablation study.} Positive values favor CGF. Exact one-sided sign tests use one paired difference per model pair; Holm correction covers the eight comparisons.}
\label{tab:e2_tests}
\end{minipage}

\clearpage
\FloatBarrier

\section{Per-Model Pair Results}

    \begin{minipage}{\textwidth}
      \centering
      \includegraphics[width=0.95\textwidth]{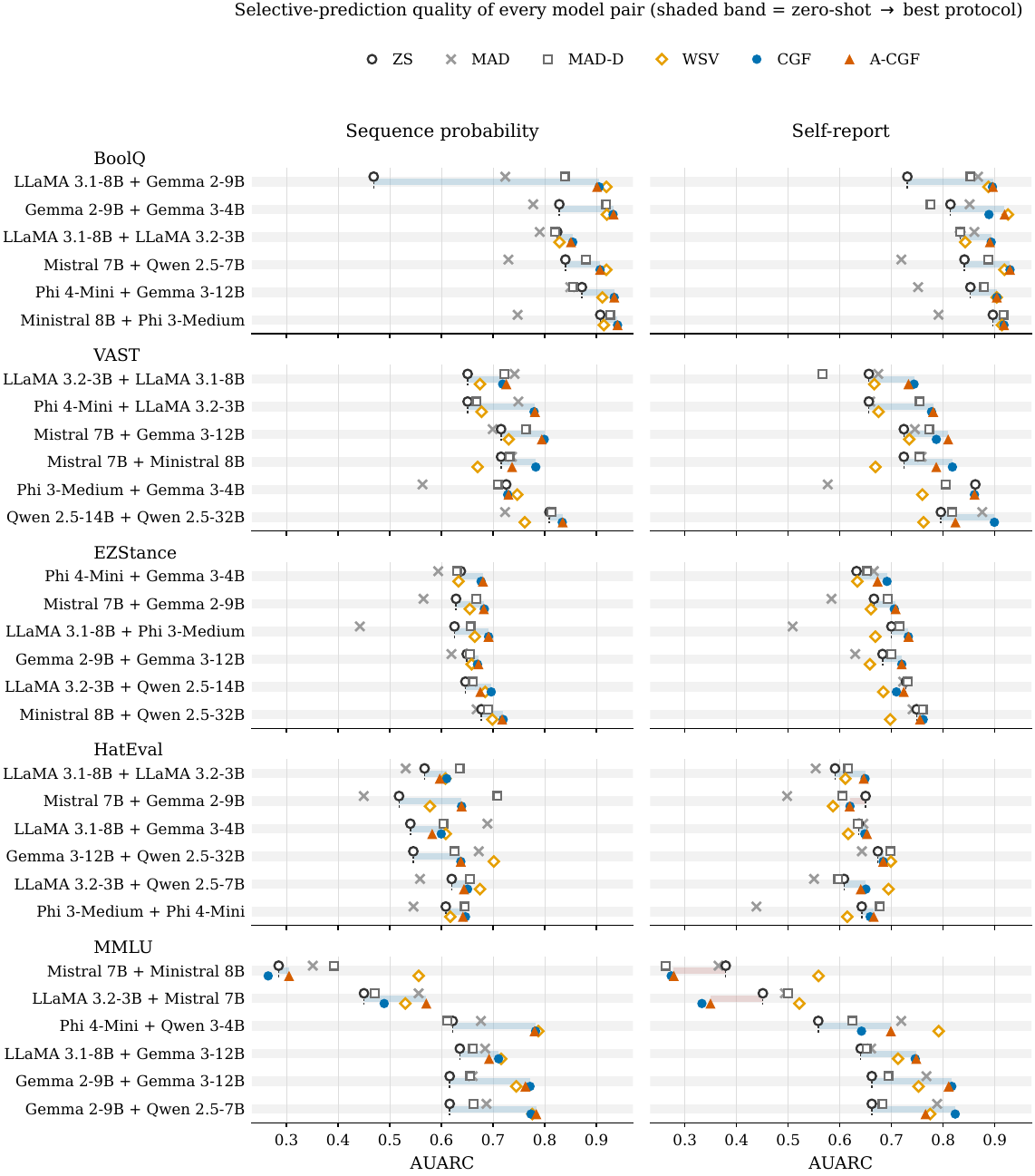}
      \captionof{figure}{\textbf{Per-model-pair AUARC.} Each row is one model pair within a dataset, shown separately for the two confidence estimators. Markers show the absolute AUARC of each baseline and
      confidence-composition protocol. The shaded segment shows the change from zero-shot to the stronger gated-fusion protocol, making the pair-level source of the dataset averages visible.}
      \label{fig:auarc_ladder}
  \end{minipage}
  \clearpage
  \begin{minipage}{\textwidth}
      \centering
      \includegraphics[width=0.95\textwidth]{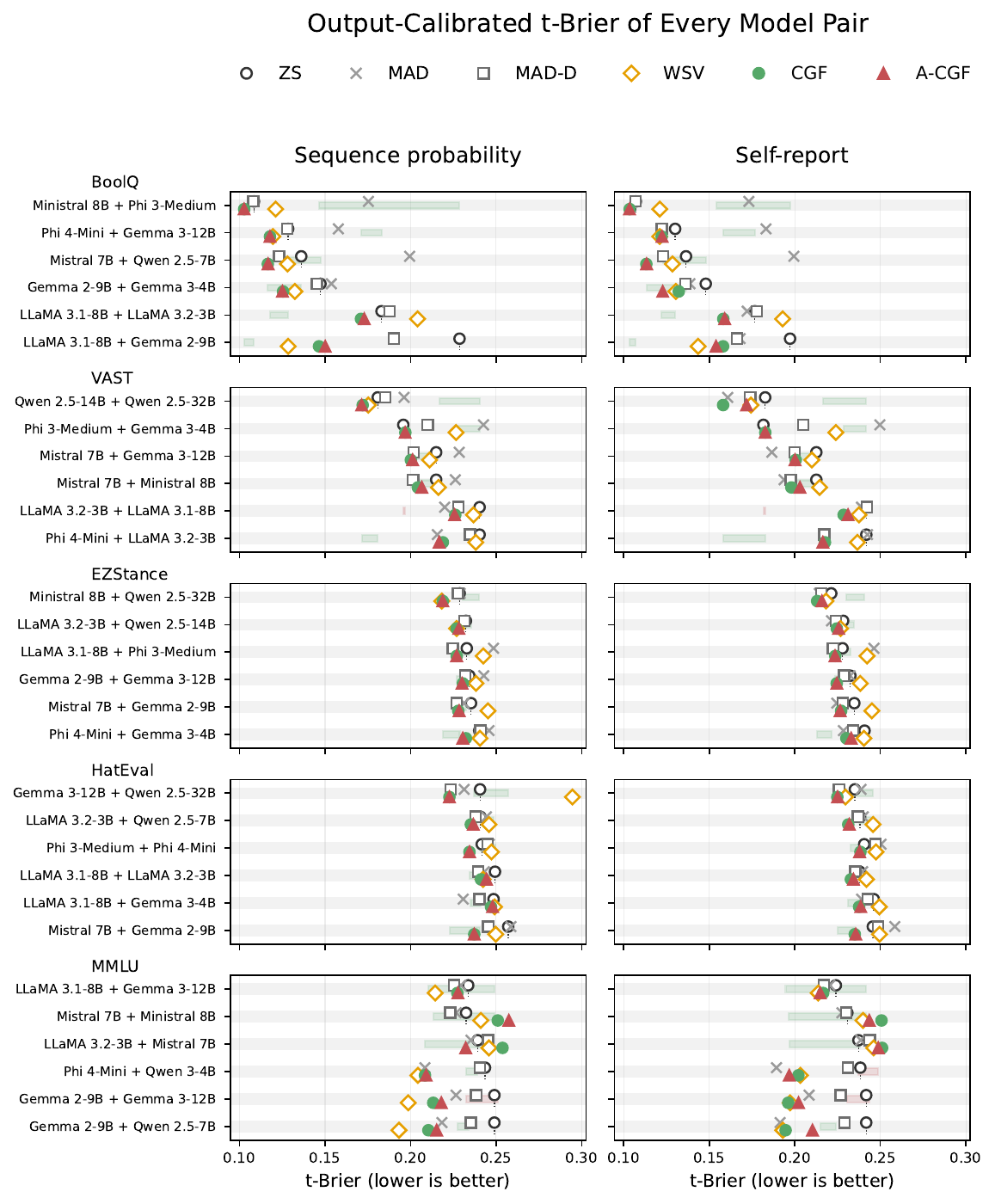}
      \captionof{figure}{\textbf{Per-model-pair t-Brier.} Each row is one model pair within a dataset, shown separately for the two confidence estimators. Markers show the temperature-scaled Brier score of each
      system, where lower values are better. The figure shows that the aggregate t-Brier gains are not driven by a single dataset or pair.}
      \label{fig:tbrier_ladder}
  \end{minipage}

  \clearpage

\clearpage
\FloatBarrier
\section{Equivalent equations for Logit-Space Fusion.}
\label{sec:eq_proof}
Let $c_1, c_2 \in (0, 1)$ represent the confidences of two model streams. Recall that the logit and logistic sigmoid functions are defined as:
\begin{align*}
\operatorname{logit}(x) &= \log\left(\frac{x}{1-x}\right) \quad \text{for } x \in (0, 1) \\
\sigma(z) &= \frac{1}{1 + e^{-z}} \quad \text{for } z \in \mathbb{R}
\end{align*}

\paragraph{Proof 1: Agreement Fusion Equivalence.}
We show that summing confidences in logit space under sigmoid activation is equivalent 
to independent Bayesian updates under agreement:

\begin{align*}
&\sigma\!\left(\operatorname{logit}(c_1) + \operatorname{logit}(c_2)\right) \\
&= \sigma\!\left( \log\left(\frac{c_1}{1-c_1}\right) + \log\left(\frac{c_2}{1-c_2}\right) \right) \\
&= \sigma\!\left( \log\left( \frac{c_1 c_2}{(1-c_1)(1-c_2)} \right) \right) \\
&= \frac{1}{1 + e^{-\log\left( \frac{c_1 c_2}{(1-c_1)(1-c_2)} \right)}} \\
&= \frac{1}{1 + \frac{(1-c_1)(1-c_2)}{c_1 c_2}} \\
&= \frac{c_1 c_2}{c_1 c_2 + (1-c_1)(1-c_2)}. \quad
\end{align*}

\paragraph{Proof 2: Disagreement Fusion Equivalence.}
We show that subtracting confidences in logit space under sigmoid activation is equivalent to the conditional probability that Stream 1 is correct given disagreement:
\begin{align*}
&\sigma\!\left(\operatorname{logit}(c_1) - \operatorname{logit}(c_2)\right) \\
&= \sigma\!\left( \log\left(\frac{c_1}{1-c_1}\right) - \log\left(\frac{c_2}{1-c_2}\right) \right) \\
&= \sigma\!\left( \log\left( \frac{c_1(1-c_2)}{c_2(1-c_1)} \right) \right) \\
&= \frac{1}{1 + e^{-\log\left( \frac{c_1(1-c_2)}{c_2(1-c_1)} \right)}} \\
&= \frac{1}{1 + \frac{c_2(1-c_1)}{c_1(1-c_2)}} \\
&= \frac{c_1(1-c_2)}{c_1(1-c_2) + c_2(1-c_1)}. \quad
\end{align*}

\section{Implementation of Multiagent Debate Baselines}
\label{sec:f1baseline}
\paragraph{ConfMAD-final.} ConfMAD \citep{lin2025enhancing} exposes each agent's calibrated confidence \emph{during} a multi-round debate and then selects the final answer of the highest-confidence agent, breaking ties uniformly at random. Our stored one-round generations cannot reproduce the in-debate intervention, since that would require re-running inference with confidence-augmented prompts. We therefore implement only the final-decision component, applied to the available post-debate outputs: the system returns the answer and confidence of whichever agent reports higher post-debate confidence. Ties are broken by a fixed coin flip derived from a hash of the input file, split and example index, which is uniform but reproducible across runs. This is an adapted control for the selection rule, not a reproduction of ConfMAD.

\paragraph{DebUnc-2agent-vote.} DebUnc \citep{yoffe2025debunc} propagates uncertainty through every debate round, either in the prompt or through modified attention, and takes an unweighted majority vote after the final round. Neither mechanism can be recreated from stored outputs, and with only two agents there is no strict majority whenever they disagree. Rather than inventing a confidence-weighted selector that DebUnc does not specify, we resolve disagreements with the same fixed, unbiased coin flip. Its reported confidence is the mean of the two post-debate confidences, used solely so that a system-level confidence can be evaluated; DebUnc itself does not define one. This adaptation is deliberately faithful rather than favourable: on datasets where the two agents frequently disagree, a coin flip on every disagreement is what produces the large F1 losses in Table~\ref{tab:f1_safeguard}.

\section{Prompts}
\label{sec:prompts}

\onecolumn

\begin{tcolorbox}[
    breakable,
    colback=white,
    colframe=black,
    title=MMLU Dataset,
    fonttitle=\bfseries\large,
    left=2mm,
    right=2mm,
    top=2mm,
    bottom=2mm
]

\small

\begin{tcolorbox}[
    breakable,
    colback=gray!5,
    colframe=black!60,
    title=Zero-shot,
    fonttitle=\bfseries,
    left=1mm,
    right=1mm,
    top=1mm,
    bottom=1mm
]

{\ttfamily\small
Provide your answer to the following question: \pur{\{Question\}}

Response: \pur{\{Agent's Response\}}
}

\end{tcolorbox}

\begin{tcolorbox}[
    breakable,
    colback=gray!5,
    colframe=black!60,
    title=Zero-shot Chain of Thought,
    fonttitle=\bfseries,
    left=1mm,
    right=1mm,
    top=1mm,
    bottom=1mm
]

{\ttfamily\small
Provide your answer to the following question: \pur{\{Question\}}

Response: \pur{\{Agent's Response\}}

Now provide a short explanation for your answer. 

Response: \pur{\{Agent's CoT\}}
}

\end{tcolorbox}

\begin{tcolorbox}[
    breakable,
    colback=gray!5,
    colframe=black!60,
    title=Debate,
    fonttitle=\bfseries,
    left=1mm,
    right=1mm,
    top=1mm,
    bottom=1mm
]

{\ttfamily\small
You are debating with another agent about the answer to the following question: \pur{\{Question\}}

Your response: \pur{\{Agent's Response\}}

Your chain-of-thought: \pur{\{Agent's CoT\}}

Other agent response: \pur{\{Other Agent's Response\}}

Other agent's chain-of-thought: \pur{\{Other Agent's CoT\}}

Evaluate the other agent's reasoning carefully and decide whether to keep your original answer, adopt theirs, or propose a new one.

Response: \pur{\{Response\}}
}

\end{tcolorbox}

\begin{tcolorbox}[
    breakable,
    colback=gray!5,
    colframe=black!60,
    title=Debate Confidence,
    fonttitle=\bfseries,
    left=1mm,
    right=1mm,
    top=1mm,
    bottom=1mm
]

{\ttfamily\small
You are debating with another agent about the answer to the following question: \pur{\{Question\}}

Your response: \pur{\{Agent's Response\}} 

Your chain-of-thought: \pur{\{Agent's CoT\}}

Other agent response: \pur{\{Other Agent's Response\}}

Other agent's chain-of-thought: \pur{\{Other Agent's CoT\}}

Evaluate the other agent’s reasoning carefully and decide whether to keep your original answer, adopt theirs, or propose a new one.

Provide your final short answer.

Your response: \pur{\{Response\}}

Rate your confidence in this final answer on a scale from 0 to 10, where 0 means you are guessing randomly, 5 means you are somewhat confident but unsure, and 10 means you are almost certain it is correct. 

Confidence: \pur{\{Confidence\}}
}

\end{tcolorbox}

\begin{tcolorbox}[
    breakable,
    colback=gray!5,
    colframe=black!60,
    title=Zero-shot Confidence,
    fonttitle=\bfseries,
    left=1mm,
    right=1mm,
    top=1mm,
    bottom=1mm
]

{\ttfamily\small
Provide your answer to the following question: \pur{\{Question\}}

Your response: \pur{\{Agent's Response\}}

Rate your confidence in this final answer on a scale from 0 to 10 - where 0 means you are guessing randomly, 5 means you are somewhat confident but unsure, and 10 means you are almost certain it is correct.

Confidence: \pur{\{Confidence\}}
}

\end{tcolorbox}

\end{tcolorbox}

\onecolumn

\begin{tcolorbox}[
    breakable,
    colback=white,
    colframe=black,
    title=Stance Detection,
    fonttitle=\bfseries\large,
    left=2mm,
    right=2mm,
    top=2mm,
    bottom=2mm
]

\small

\begin{tcolorbox}[
    breakable,
    colback=gray!5,
    colframe=black!60,
    title=Zero-shot,
    fonttitle=\bfseries,
    left=1mm,
    right=1mm,
    top=1mm,
    bottom=1mm
]

{\ttfamily\small
Analyze the stance of the author toward the topic: \pur{\{Topic\}}

The stance can fall into one of three categories: Against, Supporting, or Neutral.

Sentence: \pur{\{Sentence\}}

Detect the stance of the writer toward. Give only one word as a response with \{Neutral, Support, Against\}.

Response: \pur{\{Agent's Response\}}
}

\end{tcolorbox}

\begin{tcolorbox}[
    breakable,
    colback=gray!5,
    colframe=black!60,
    title=Zero-shot Chain of Thought,
    fonttitle=\bfseries,
    left=1mm,
    right=1mm,
    top=1mm,
    bottom=1mm
]

{\ttfamily\small
Analyze the stance of the author toward the topic: \pur{\{Topic\}}

The stance can fall into one of three categories: Against, Supporting, or Neutral.

Sentence: \pur{\{Sentence\}}

Detect the stance of the writer toward. Give only one word as a response from \{Neutral, Support, Against\}.

Response: \pur{\{Agent's Response\}}

Now provide a short explanation for your answer \pur{\{Agent's CoT\}}

Response: \pur{\{Response\}}
}

\end{tcolorbox}

\begin{tcolorbox}[
    breakable,
    colback=gray!5,
    colframe=black!60,
    title=Debate,
    fonttitle=\bfseries,
    left=1mm,
    right=1mm,
    top=1mm,
    bottom=1mm
]

{\ttfamily\small
You are debating with another agent about the stance of the following sentence toward the topic: 
\pur{\{Topic\}}

Sentence: \pur{\{Sentence\}}

Your response: \pur{\{Agent's Response\}}

Your chain of thought: \pur{\{Agent's CoT\}}

Other agent response: \pur{\{Other Agent's Response\}}

Other agent's chain of thought:  \pur{\{Other Agent's CoT\}}

Critically analyze the other agent's reasoning. Maintain your analytical perspective. Provide your updated answer. Only respond with one word from \{Neutral, Support, Against\}.

Response:  \pur{\{Response\}}
}

\end{tcolorbox}

\begin{tcolorbox}[
    breakable,
    colback=gray!5,
    colframe=black!60,
    title=Debate Confidence,
    fonttitle=\bfseries,
    left=1mm,
    right=1mm,
    top=1mm,
    bottom=1mm
]

{\ttfamily\small
You are debating with another agent about the stance of the following sentence toward the topic: 
 \pur{\{Topic\}}

Sentence: \pur{\{Sentence\}}

Your response: \pur{\{Agent's Response\}} 

Your chain of thought: \pur{\{Agent's CoT\}}

Other agent response: \pur{\{Other Agent's Response\}} 

Other agent's chain of thought: \pur{\{Other Agent's CoT\}} 

Critically analyze the other agent's reasoning. Maintain your analytical perspective. Provide your updated answer. Only respond with one word from \{Neutral, Support, Against\}.

Response: \pur{\{Response\}} 

Now report your confidence level for your answer out of 10. Only output one integer in the [0, 10] range.

Confidence: \pur{\{Confidence\}} 
}

\end{tcolorbox}

\begin{tcolorbox}[
    breakable,
    colback=gray!5,
    colframe=black!60,
    title=Zero-shot Confidence,
    fonttitle=\bfseries,
    left=1mm,
    right=1mm,
    top=1mm,
    bottom=1mm
]

{\ttfamily\small
Analyze the stance of the author toward the topic: \pur{\{Topic\}} 

The stance can fall into one of three categories: Against, Supporting, or Neutral.

Sentence: \pur{\{Sentence\}} 

Detect the stance of the writer toward. Give only one word from \{Neutral, Support, Against\}.

Response: \pur{\{Response\}} 

Now report your confidence level for your answer out of 10. Only output one integer in the [0, 10] range.

Confidence: \pur{\{Confidence\}} 
}

\end{tcolorbox}

\end{tcolorbox}


\onecolumn

\begin{tcolorbox}[
    breakable,
    colback=white,
    colframe=black,
    title=Logical Question Answering,
    fonttitle=\bfseries\large,
    left=2mm,
    right=2mm,
    top=2mm,
    bottom=2mm
]

\small

\begin{tcolorbox}[
    breakable,
    colback=gray!5,
    colframe=black!60,
    title=Zero-shot,
    fonttitle=\bfseries,
    left=1mm,
    right=1mm,
    top=1mm,
    bottom=1mm
]

{\ttfamily\small
You will receive a context and a statement/question. Your task is to determine if the statement is true or false based on the context.

Context: \pur{\{Context\}} 

Statement/Question: \pur{\{Statmenst/Question\}} 

Choose only one word for response from \{true, false\}.

Your response: \pur{\{Agent's Response\}} 
}

\end{tcolorbox}

\begin{tcolorbox}[
    breakable,
    colback=gray!5,
    colframe=black!60,
    title=Zero-shot Chain of Thought,
    fonttitle=\bfseries,
    left=1mm,
    right=1mm,
    top=1mm,
    bottom=1mm
]

{\ttfamily\small
You will receive a context and a statement/question. Your task is to determine if the statement is true or false based on the context.

Context: \pur{\{Context\}}

Statement/Question: \pur{\{Statement/Question\}}

Choose only one word for response from \{true, false\}.

Your response: \pur{\{Agent's Response\}}

Now provide a short explanation for your answer \pur{\{Agent's CoT\}}

Response: \pur{\{Response\}}
}

\end{tcolorbox}

\begin{tcolorbox}[
    breakable,
    colback=gray!5,
    colframe=black!60,
    title=Debate,
    fonttitle=\bfseries,
    left=1mm,
    right=1mm,
    top=1mm,
    bottom=1mm
]

{\ttfamily\small
You and another agent are debating whether the answer to a question is True or False, using a given context as evidence.

Context: \pur{\{Context\}}

Statement/Question: \pur{\{Statement/Question\}}

Your response: \pur{\{Agent's Response\}}

Your chain-of-thought: \pur{\{Agent's CoT\}}

Other agent response: \pur{\{Other Agent's Response\}}

Other agent's chain-of-thought: \pur{\{Other Agent's CoT\}}

Critically analyze the other agent's reasoning. Maintain your analytical perspective.

Provide your updated answer. Choose only one word from \{true, false\}.

Response: \pur{\{Response\}}
}

\end{tcolorbox}

\begin{tcolorbox}[
    breakable,
    colback=gray!5,
    colframe=black!60,
    title=Debate Confidence,
    fonttitle=\bfseries,
    left=1mm,
    right=1mm,
    top=1mm,
    bottom=1mm
]

{\ttfamily\small
You and another agent are debating whether the answer to a question is True or False, using a given context as evidence.

Context: \pur{\{Context\}}

Statement/Question: \pur{\{Statement/Question\}}

Your response: \pur{\{Agent's Response\}}

Your chain-of-thought: \pur{\{Agent's CoT\}}

Other agent response: \pur{\{Other Agent's Response\}}

Other agent's chain-of-thought: \pur{\{Other Agent's CoT\}}

Critically analyze the other agent's reasoning. Maintain your analytical perspective.

Provide your updated answer. Choose only one word from \{true, false\}.

Response: \pur{\{Response\}}

Now report your confidence level for your answer out of 10. Only output one integer in the [0, 10] range.

Your Confidence: \pur{\{Confidance\}}
}

\end{tcolorbox}

\begin{tcolorbox}[
    breakable,
    colback=gray!5,
    colframe=black!60,
    title=Zero-shot Confidence,
    fonttitle=\bfseries,
    left=1mm,
    right=1mm,
    top=1mm,
    bottom=1mm
]

{\ttfamily\small
You will receive a context and a statement/question. Your task is to determine if the statement is true or false based on the context.

Context: \pur{\{Context\}}

Statement/Question: \pur{\{Statement/Question\}}

Choose only one word from \{true, false\}.

Your response: \pur{\{Agent's Response\}}

Now report your confidence level for your answer out of 10. Only output one integer in the [0, 10] range.

Your Confidence: \pur{\{Confidence\}}
}

\end{tcolorbox}

\end{tcolorbox}


\onecolumn

\begin{tcolorbox}[
    breakable,
    colback=white,
    colframe=black,
    title=Hate Speech Detection,
    fonttitle=\bfseries\large,
    left=2mm,
    right=2mm,
    top=2mm,
    bottom=2mm
]

\small

\begin{tcolorbox}[
    breakable,
    colback=gray!5,
    colframe=black!60,
    title=Zero-shot,
    fonttitle=\bfseries,
    left=1mm,
    right=1mm,
    top=1mm,
    bottom=1mm
]

{\ttfamily\small
Evaluate the following sentence, return true if the sentence contains hate speech, otherwise return false.

Choose only one word for response from \{true, false\}.

Response: \pur{\{Agent's Response\}}
}

\end{tcolorbox}

\begin{tcolorbox}[
    breakable,
    colback=gray!5,
    colframe=black!60,
    title=Zero-shot Chain of Thought,
    fonttitle=\bfseries,
    left=1mm,
    right=1mm,
    top=1mm,
    bottom=1mm
]

{\ttfamily\small
Evaluate the following sentence, return true if the sentence contains hate speech, otherwise return false.

Choose only one word for response from \{true, false\}.

Response: \pur{\{Agent's Response\}}

Now provide a short explanation for your answer \pur{\{Agent's CoT\}}

Response: \pur{\{Response\}}
}

\end{tcolorbox}

\begin{tcolorbox}[
    breakable,
    colback=gray!5,
    colframe=black!60,
    title=Debate,
    fonttitle=\bfseries,
    left=1mm,
    right=1mm,
    top=1mm,
    bottom=1mm
]

{\ttfamily\small
You are debating with another agent about whether the following sentence contains hate speech \pur{\{sentence\}}

Your response: \pur{\{Agent's Response\}}

Your chain of thought: \pur{\{Agent's CoT\}}

Other agent response: \pur{\{Other Agent's Response\}}

Other agent's chain of thought: \pur{\{Other Agent's CoT\}}

Critically analyze the other agent's reasoning. Maintain your analytical perspective.

Provide your updated answer. Respond with only one word from \{true, false\}.

Response: \pur{\{Response\}}
}

\end{tcolorbox}

\begin{tcolorbox}[
    breakable,
    colback=gray!5,
    colframe=black!60,
    title=Debate Confidence,
    fonttitle=\bfseries,
    left=1mm,
    right=1mm,
    top=1mm,
    bottom=1mm
]

{\ttfamily\small
You are debating with another agent about whether the following sentence contains hate speech: \pur{\{Sentence\}}

Your response: \pur{\{Agent's Response\}}

Your chain of thought: \pur{\{Agent's CoT\}}

Other agent response: \pur{\{Other Agent's Response\}}

Other agent's chain of thought: \pur{\{Other Agent's CoT\}}

Critically analyze the other agent's reasoning. Maintain your analytical perspective.

Provide your updated answer. Respond with only one word from \{true, false\}.

Response: \pur{\{Response\}}

Now report your confidence level for your answer out of 10. Only output one integer in the [0, 10] range.

Confidence: \pur{\{Confidence\}}
}

\end{tcolorbox}

\begin{tcolorbox}[
    breakable,
    colback=gray!5,
    colframe=black!60,
    title=Zero-shot Confidence,
    fonttitle=\bfseries,
    left=1mm,
    right=1mm,
    top=1mm,
    bottom=1mm
]

{\ttfamily\small
Evaluate the following sentence, return true if the sentence contains hate speech, otherwise return false \pur{\{Sentence\}}

Response: \pur{\{Agent's Response\}}

Now report your confidence level for your answer out of 10. Only output one integer in the [0, 10] range.

Confidence: \pur{\{Confidence\}}
}

\end{tcolorbox}

\end{tcolorbox}

\twocolumn

\section{AI Usage}
AI is used to edit the text for this submission. It was not used to generate an entire sentence or paragraph, but to make sentences easier to read. Also, to improve the text's grammar.

For coding, AI was used for small-grain sub-function code or small utility functions, and for debugging.

\end{document}